\documentclass{article}

\usepackage[utf8]{inputenc}
\usepackage{amsmath}
\usepackage{amssymb}
\usepackage{stmaryrd}
\usepackage{graphicx}
\usepackage{float}

\DeclareMathOperator*{\argmax}{argmax}

\title{Relational Data Mining Through Extraction of Representative Exemplars}
\author{Frédéric Blanchard \footnote{frederic.blanchard@univ-reims.fr}   and Michel Herbin \footnote{michel.herbin@univ-reims.fr}}

\begin{document}

\bibliographystyle{plain} 

\maketitle

\paragraph{Abstract:~}
With the growing interest on Network Analysis, Relational Data Mining is becoming an emphasized domain of Data Mining. This paper addresses the problem of extracting representative elements from a relational dataset. After defining the notion of \emph{degree of representativeness}, computed using the Borda aggregation procedure, we present the extraction of \emph{exemplars} which are the representative elements of the dataset. We use these concepts to build a \emph{network} on the dataset. We expose the main properties of these notions and we propose two typical applications of our framework. The first application consists in resuming and structuring a set of binary images and the second in mining co-authoring relation in a research team.
\paragraph{Keywords:~}relational data, data mining, representative, exemplar, clustering, network, Borda.

\section{Introduction}

The data mining is interested in discovering knowledge from data. Nowadays finding interesting patterns or structures is a crucial task in the field of data analysis. Thus the paper addresses the problem of the extraction of representative elements from a dataset. This problem presents a significant interest when designing recommendation systems \cite{Pazzani2007}, selecting leaders or specimens \cite{Brun2010}, community detection \cite{Newman2004}, customer Relationship analysis \cite{Tuzhilin2012} or sub-sampling.\\

The classical ways to determine representative elements refer to the task of data clustering \cite{Alfred2010}. The goal is to partition of the dataset. Then the representative elements are the prototypes of the clusters. They can be chosen as the average elements of each cluster or selected after a random initialization step. For instance, when using $k-$means or $k-$centers algorithms (see \cite{Jain1999} for a review of clustering methods including $k-$means algorithm), the centers of the obtained clusters provide the prototypes of the dataset. The prototypes are first randomly selected and the algorithms iteratively refine the set of prototypes. The final elements are quite sensitive to the initial selection. Moreover $k-$means algorithm leads to average prototype which are not ``real'' elements of the initial dataset. this kind of methods is not satisfactory. The lacks of all the approaches based on clustering are multiple. Firstly the partition into clusters is predate to the extraction of representative elements and the clusters have to be validated and interpreted to justify the prototypes. Secondly the representative elements depend on the choice of clustering algorithms and the extraction of the prototypes depends on the implicit assumptions about the shape of clusters and data distributions. Moreover when one cluster contains more than one sub-population, only one prototype is extracted. Finally, in the case of clustering algorithms like $k$-means, the centers are not elements of the original dataset. They are average computed elements.  How make a mean-element meaningful ? Most of the time, providing a non-existing element (virtual element like a mean-element) does not make sense.\\

In this paper the approach we present consists in extracting elements we called \emph{exemplars} directly from the whole dataset, without any a priori clustering step (in one pass unlike \cite{Frey2007}) . The exemplars summarize the dataset and are particular elements of the original dataset. Thus they are real data. These elements are as representative as possible of the whole set without any assumption on the shape or the density of data distribution (unlike in \cite{Luhr2008}).
To achieve the extraction of exemplar, we construct a \emph{degree of representativeness} on the dataset. The exemplars are finally chosen as local maxima of the degree of representativeness. By fitting the locality parameter (in topological terms the \emph{scale factor}) we adapt the scale to determine the number of exemplars.\\

The paper is organized as follows. In the first section we introduce the context and expose our method. We present the formal definitions of \emph{scores} between data, the notion of \emph{standard} and the concept of \emph{exemplars} in the dataset. Then we show how to build a network of exemplars to visualize these notions. For each definition we present some interesting and remarkable properties (robustness, stability etc.) \\
In Section 3, we provide two applications in very different context. Firstly we apply our method on a set of binary images. We compute scores and exemplars and we build the network to structure the dataset. The second application concerns the analysis of co-authoring in a research laboratory. We exhibit a co-authoring network that permits to visualize how researchers are really clustered and how they work together.\\
Section 4 is a brief conclusion that outlines our main contributions and that expose our current and future works.

\section{Method}

Let $\Omega$ be a set of $n$ elements 
in a multidimensional space.
The $n$ elements are qualitative, quantitative or mixed data.
We assume that $\Omega$ is a relational data set
without any underlying distribution.
Let us describe the way we use to extract the exemplars of $\Omega$ structuring this set in a network.
In this paper, the elements are called objects.

\subsection{Pairwise Valued Relation}

$\Omega$ is a relation dataset.
Let us specify this relation.
Let $R$ be a pairwise valued relation on $\Omega$. 
$R$ is defined by :
\begin{displaymath}
\begin{array}{rrcl}
	R: & \Omega \times \Omega &\rightarrow &\mathbb{R}^+ \\
	                     & (x,y) & \mapsto & R(x, y)
\end{array}
\end{displaymath}
\\
The use of a pairwise valued relation is 
very useful in data processing.
A distance is a special case of this kind of relation.
But a distance is frequently not available 
when processing qualitative data.
Thus a relation is more widespread than a distance
for pairwise comparisons of objects.
In this paper, the value $R(x,y)$ is also called 
the \emph{cost} from $x$ to $y$,
indicating the generality of the relation.
\\
The relation must follow three trivial properties.
\begin{itemize}
\item The relation must be \emph{total}.
This means that each pair of objects of $\Omega$ is valued by $R$.
\item The relation must be \emph{positive}. 
The cost is a positive value for all pairs.
\item The cost from $x$ to $x$ is null forall $x$ (i.e. $\forall x \in \Omega, R(x,x)=0$)
\end{itemize}

Unlike a distance, the relation does not necessarily respect
the property of symmetry.
$R(x, y)$ may be different from $R(y, x)$.
For instance, if the cost from a point $x$ to a point $y$ 
is the time to go from $x$ to $y$, 
then the cost from $y$ to $x$ could differ from the first one
because of the slope, wind, flow, etc.
Moreover, the relation does not respect the triangle inequality.
A dissimilarity index gives a classical example of such a relation
which does not respect the triangle inequality.
$x$ is dissimilar from $y$ with $R(x, y)$ and 
$y$ is dissimilar from $z$ with $R(y, z)$
but $x$ could be dissimilar from $z$ with $R(x, z) > R(x, y) + R(y, z)$.
\\
Such a relation can lead to a vote 
to designate exemplars within the dataset.
Specifically, we can rank the objets of $\Omega$
taking into account the relation
to set up votes between the objects themselves.
The following subsection describes this procedure.

\subsection{Score}

\begin{figure}[H]
\center
\includegraphics[width=1\textwidth,height=0.5\textwidth]{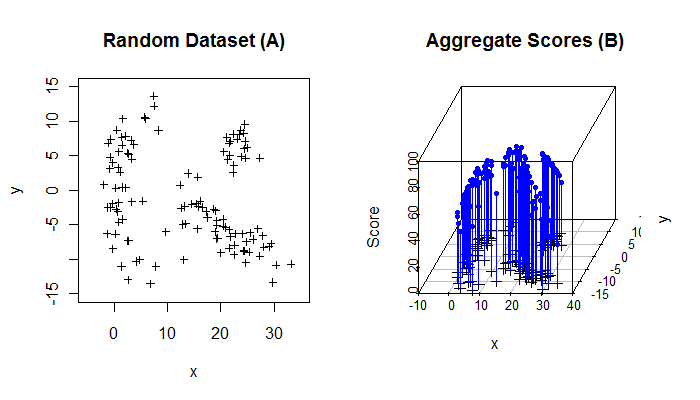}
\caption{Example of a dataset with 120 random samples (A) and their respective aggregated scores (B).
The score increases in the midst of the dataset}
\label{dataSetScores}
\end{figure}

In this paper, we select an exemplar object from $\Omega$
according to the Borda voting method
\cite{deBorda1781}. But firstly, we transform values of the relation into ranks \cite{Barnett1976}\cite{Conover1981}\cite{David2003}.
Let $x$ be an object of $\Omega$. 
All objects can be sorted 
by the ascending order of their costs relative to $x$. 
Let us note $Rk_{x}(y)$ the rank of  $y$ relative to $x$.
The rank is obtained when sorting the set $\{R(x,z)/z\in \Omega\}$. 
Using Borda method \cite{deBorda1781}\cite{vanErp2000}, the object $x$ assigns a relative score 
to all objects of $\Omega$. 
The score $Sc_{x}$ relative to $x$ is defined by:
\begin{displaymath}
\forall y\in \Omega, Sc_{x}(y) = n - Rk_{x}(y)
\end{displaymath}
where $n$ is the number of objects of $\Omega$.
Thereby the relative score is an integer 
and it lies between 0 and $n-1$.
The lower the cost from $x$ to $y$,
the higher the score of $y$ relative to $x$.
\\
Computing all relative scores, 
each object $x$ receives $n$ relative scores
corresponding to the votes of all objects of $\Omega$
(i.e. the $n$ values $Sc_{y}(x)$ with $y \in \Omega$). 
Then the relative scores are aggregated 
to choose the winner of the voting procedure.
The aggregate score is defined by:
\begin{displaymath}
\begin{array}{rrcl}
	Sc: & \Omega &\rightarrow &\mathbb{R}^+ \\
	    & x & \mapsto & Sc(x) = \displaystyle \frac{1}{n}\sum_{y\in \Omega}Sc_{y}(x)
\end{array}
\end{displaymath}
In this paper, the aggregation function is the $mean$ function.
\\
Let us observe the aggregated scores in a relational dataset.
Figure~\ref{dataSetScores} displays an example of a dataset 
with 120 two dimensional random samples (A).
Euclidean distance is used as the pairwise valued relation between samples.
The respective aggregated scores (B) confirm that the score increases 
when the sample approaches the center of the dataset,
i.e. in the midst of this one.

\subsection{Standard}

\begin{figure}[H]
\center
\includegraphics[width=1\textwidth,height=0.4\textwidth]{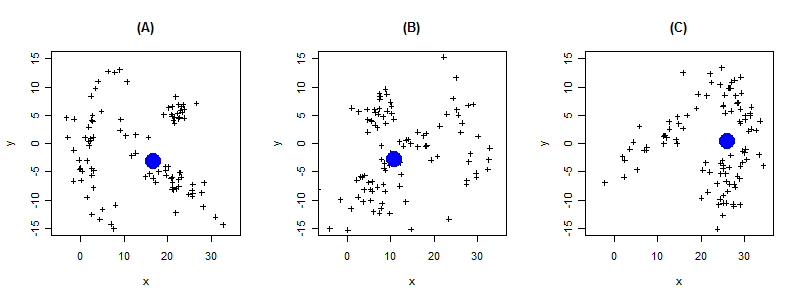}
\caption{Standard examples (filled circles) 
for respectively the datasets (A), (B), and (C).
The datasets have 100 random samples. 
The aggregated scores of the standards are respectively 68.75, 70.55, and 68.77. 
}
\label{StandardExemples}
\end{figure}

The object with the highest aggregated score
is the \emph{standard} we propose.
Let us observe some properties of the \emph{standard}.
\\
Figure~\ref{StandardExemples} displays three datasets A, B, and C.
Each dataset has 100 random samples ($n = 100$).
The aggregated scores are computed using Euclidean distance as pairwise valued relation.
The maxima of aggregated score are respectively 68.75, 70.55, and 68.77 for A, B and C.
Filled circles indicate the three respective standards with the highest aggregated scores.
Figure~\ref{StandardExemples} confirms that each standard lies in the midst of its dataset.

\begin{figure}[H]
\center
\includegraphics[width=1\textwidth,height=0.4\textwidth]{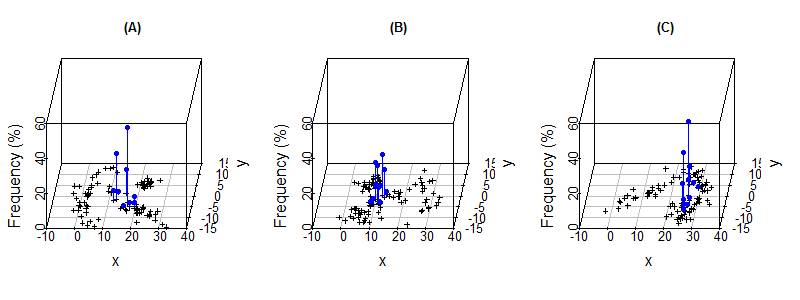}
\caption{Frequencies of the standards when resampling the datasets (A), (B), and (C) of Figure~\ref{StandardExemples}.
The crosses show the 100 initial objects of the datasets.
Stems with filled circles show the frequencies of the standards obtained when using 200 bootstraps. The most frequent standards appear in (A), (B), and (C) with respectively the frequencies of 40\%, 32\%, 36\%.
}
\label{StandardFrequencies}
\end{figure}

When resampling the dataset using the bootstrap technique \cite{Thomas2000}, 
the standard could change.
If it does not change, the extraction of this standard is robust against the resampling.
Using many bootstraps, the highest frequency of the extracted standards indicates the stability of the standard when resampling.
Our experiments using simulated data and real data show that the standard depends very weakly in the resampling.
\\
Figure~\ref{StandardFrequencies} displays the standards 
obtained when resampling the datasets (A), (B), and (C) of Figure~\ref{StandardExemples}.
The initial datasets have 100 elements displayed with crosses.
Stems with filled circles show the frequencies of the standards obtained with 200 bootstraps.
The extracted standards remain in the center of respectively A, B, and C.
The frequencies of the most frequent standards when resampling the 100 initial samples 
are respectively equal to 40\%, 32\%, 36\%.
These frequencies  assess the stability of the standard with respect to the samples.
Respectively 90\%, 88\%, and 90\% of the dataset elements are never extracted as standards when resampling.
\\
Thus we assume that a standard gives a clue on the center of the dataset.
Because the standard is a real element, it avoids the nonsense 
that the classical averages could produce with a virtual out-of-scope element
outside of the data distribution.
Note that the stability of the standard 
(i.e. the frequency of the most frequent standard)
increases when the number of objects increases.

Let us examine the stability of the standard when outliers are feared.
We simulate outliers that we append to an initial dataset.
We consider that the standard extraction is robust against outliers 
when the extracted standard remains one of 
three most frequent standards of the initial dataset.
\\
In this paper we describe the study of robustness (see \cite{Rousseeuw2003} for more details about the concept of robustness)
using the datasets A, B and C of Figure~\ref{StandardExemples}.
The outliers are random elements out of the range of 
the initial data domain.
In this section, the domain is defined by elements of coordinates $(x, y)$
where $-10 \le x \le 40$ and $-15 \le y \le 15$.
Outliers are simulated in a larger domain 
defined by $-10000 \le x \le 40000$ and $-15000 \le y \le 15000$
(the initial limits are multiplied by 1000)
excluding the elements that are too close from the initial domain
by keeping the elements $(x, y)$ where
$x \le -1000$ or $4000 \le x$ and $y \le -1500$ or $1500 \le y$
(the limits of initial domain are multiplied by 100).
We add such random outliers to an initial dataset
until the extracted standards changes 
(i.e. until the extracted standard from the new dataset with outliers 
will not be one of the three most frequent standards of the initial dataset).
When outliers are randomly generated in a such very large domain,
the percentage of outliers could be higher than 200\% 
without changing the initial standard.
Then the standard is robust when the outliers are spread in a large domain. 
But the standard remains also robust when outliers are concentrated into only one duplicate object. 
When only one outlier is randomly generated in the very large domain,
we could add up to 20\% of out-of-range elements using 
this single outlier without changing the initial standard.
Then we assume that the standard is particularly robust against outliers.

\subsection{Exemplars and Networks}

\begin{figure}[H]
\center
\includegraphics[width=1\textwidth,height=1\textwidth]{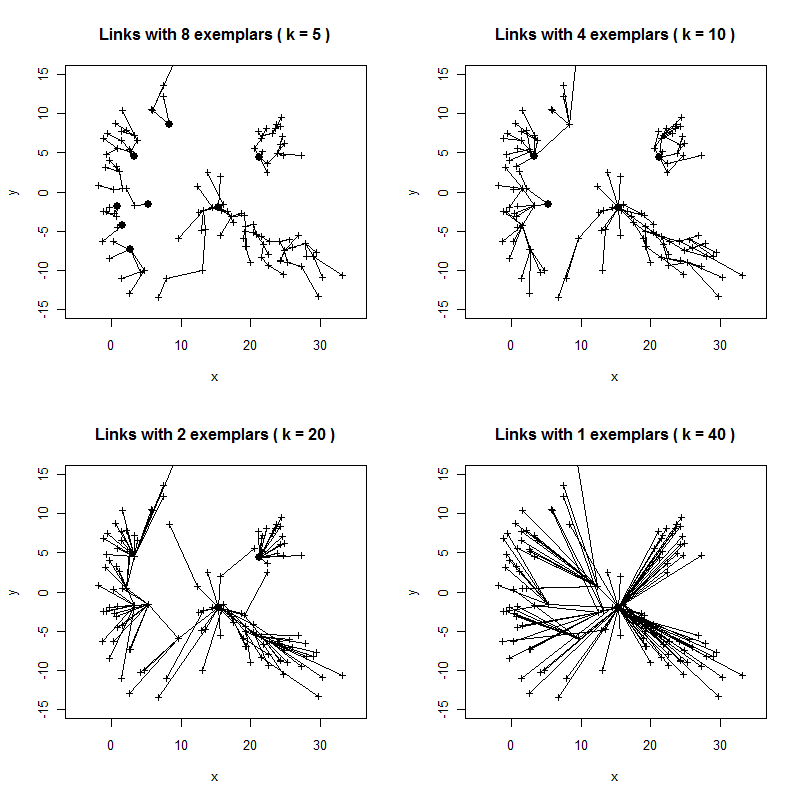}
\caption{Networks obtained with scale factor $k=5$, $k=10$, $k=20$, and $k=40$.
The networks are built between the 120 samples of Figure~\ref{dataSetScores} (A).
The exemplars are displayed with black filled circles.
}
\label{networkScaleFactor}
\end{figure}

The standard is the only exemplar extracted from a dataset.
But the dataset may be complex and 
it could require more than one exemplar 
to represent the whole set.
This section describes how the dataset can be structured 
to retrieve these exemplars from the set.
\\
The first step consists in defining the neighborhood of each object within $\Omega$. 
Let $x$ be one of the $n$ objects of $\Omega$.
Let $k$ be a value between 0 and $n$.
The $k$-nearest neighbors of $x$ are defined using the ranks relative to $x$.
Then the $k-$neighborhood of $x$ in $\Omega$ is defined by:
\begin{displaymath}
\forall x \in \Omega, \quad
\forall k \in \llbracket 1,n \rrbracket,  \quad
N_{k}(x) = \{y\in \Omega / Rk_{x}(y) \leq k\}
\end{displaymath}
Thus $N_{k}(x)$ is the set of $k$ nearest objects of $x$.
\\
In a second step, each object $x$ is associated with 
the neighbor having the highest aggregated score. 
Thus we define a link from $x$ to its preferred neighbor. 
Each object $x$ is linked to an object $y$.
The links are defined by:
\begin{displaymath}
\forall x\in \Omega, \quad
x \mapsto y=\argmax_{z\in N_k(x)}Sc(z)
\end{displaymath}
In this definition, $x$ is linked to $y$ and $y$ is generally different from $x$ 
when $Sc(y)> Sc(x)$. 
If $Sc(x)$ is maximal inside $N_{k}(x)$, then $y=x$ 
and $x$ is linked to $x$ itself. 
These self-linked objects are simply called \emph{exemplars} of $\Omega$.
\\
Using the links, the dataset becomes a network 
where the nodes are the objects.
The exemplars becomes the terminal nodes of this network 
(i.e. the roots of the trees forming the network).
The exemplars depend on the value of $k$
which influences the network configuration.
In this paper, $k$ is the size of the neighborhood we use.
This parameter is called scale factor.
\\
Figure~\ref{networkScaleFactor} displays four networks
obtained from the simulated dataset of Figure~\ref{dataSetScores} (A).
The dataset has the 120 samples ($n=120$).
The four networks are configured
using the scale factors 5, 10, 20, and 40.
The exemplars are displayed with a filled circle, 
they are the terminal nodes of the networks.
The numbers of extracted exemplars
are respectively equal to 8, 4, 2 and 1.
Distinctly the number of exemplars
depends on the scale factor $k$.
The following describes the influence of the scale factor.

\subsection{Exemplars and Scale Factor}

\begin{figure}[H]
\center,
\includegraphics[width=0.5\textwidth,height=0.5\textwidth]{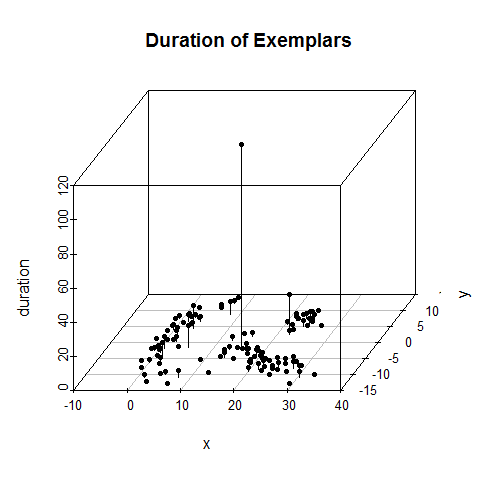}
\caption{
Duration of exemplars increasing the scale factor: The Figure~\ref{dataSetScores} dataset has 120 objects ($n=120$). 
The scale factor increases from 1 to 120. 
When the scale factor is equal to 1, all the objects are exemplars.
When the scale factor increases, some exemplars disappear. 
Only the standard is always extracted when increasing scale factor.
Then its duration is equal to 120.
}
\label{duration}
\end{figure}

The higher the scale factor, 
the lower the number of exemplars. 
Moreover, when the scale factor increases from one to $n$, 
the number of exemplars decreases from $n$ to one. 
Let us explain this property.
When $k=1$,  $N_{1}(x)$ is the singleton equal to ${x}$.
Therefore each object $x$ is itself an exemplar of $\Omega$
(i.e. $x$ is linked to $x$).
Then the set of exemplars is $\Omega$
and the number of exemplars is equal to $n$.
When $k=n$, $N_{n}(x)$ is equal to $\Omega$.
Each object $x$ is linked to the standard
which has the highest aggregated score within $\Omega$.
Then the number of exemplars is equal to $1$
the network becomes only one tree
and the standard is its root.
At the scale $k$, an exemplar $x$ has 
the highest aggregated score
within the neighborhood  $N_{k}(x)$ 
(i.e. within the $k$ nearest neighbors of $x$).
If $k_{1} \leq k_{2}$, then $N_{k_{1}}(x) \subseteq N_{k_{2}}(x)$. 
If $x$ is an exemplar at the scale $k_{2}$, 
then it is an exemplar at the scale $k_{1}$. 
Therefore the number of exemplars necessarily decreases
when the scale factor increases.
\\
Increasing the scale factor, 
some exemplars could disappear 
among those who were extracted. 
But an object never appears as an exemplar
if it was not extracted at lower scale factor.
Figure~\ref{duration} displays the duration of each exemplar when increasing the scale factor.
The exemplars are extracted from Figure~\ref{dataSetScores} dataset ($n=120$).
When the scale factor is equal to 1, all the objects are exemplars.
When the scale factor increases, some exemplars disappear and
their duration is shortened.
Only the standard is kept from scale 1 to the scale $n$.
It has the longest duration equal to $n$.

\begin{figure}[H]
\center,
\includegraphics[width=0.8\textwidth,height=0.8\textwidth]{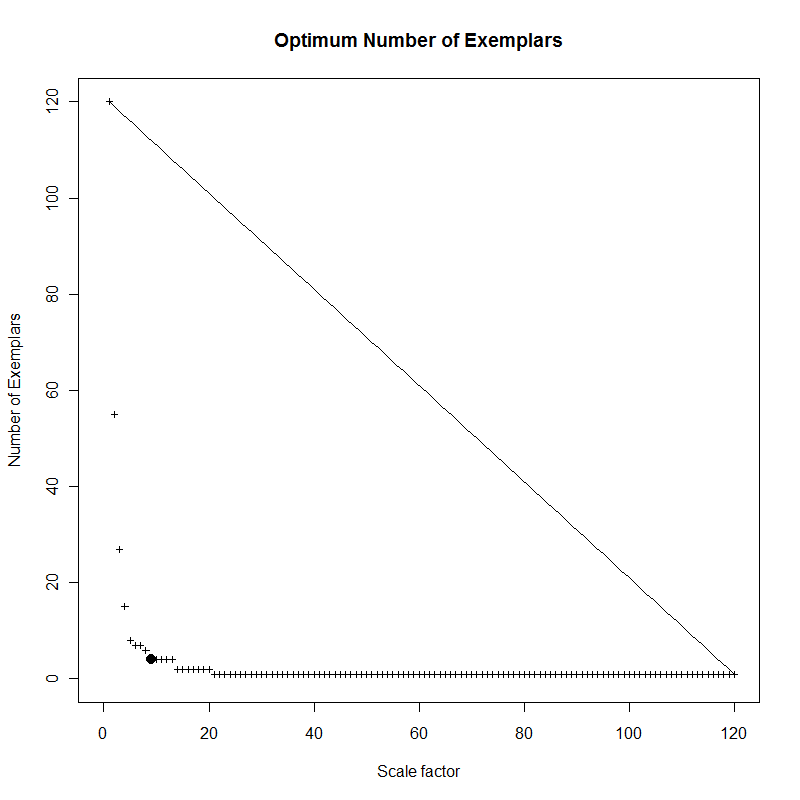}
\caption{
Number of Exemplars from Figure~\ref{dataSetScores} (A) dataset and Scale Factor : The number of exemplar is smaller than $120 - (k-1)$ where
$k$ is the scale factor and 
$120$ is the number of objects of the dataset.
The difference between $120 - (k-1)$ (line) and the number of extracted exemplars (cross) is maximal when the scale factor is equal to 9 (filled circle).}
\label{functionScaleFactor}
\end{figure}

At the scale $k$, we assume that the numbers of exemplars 
is smaller than $n - (k-1)$ where
$k$ is the scale factor and 
$n$ is the number of objects of the dataset.
At each scale $k$, we want to reduce the number of exemplars.
When this number is equal to $n-k+1$, 
we consider that the extraction of exemplars is suboptimal.
This case is observed when $k=1$ or $k=n$.
In this paper, the scale factor becomes optimal
when the difference between $n-k+1$
and the number of extracted exemplars is maximum.
Let $k_{optimum}$ be this optimal value
of the scale factor we propose in this paper.
\\
Figure~\ref{functionScaleFactor} displays the numbers of exemplars 
according to the scale factor $k$.
It uses the dataset of Figure~\ref{dataSetScores} (A) ($n=120$).
The scale factor increases from 1 to 120 and the number of exemplars decreases from 120 to 1.
The numbers of exemplars is smaller than $121-k$. 
The difference between $121-k$
and the number of exemplars is maximum when $k=9$.
The black filled circle shows this optimum value.
Then four exemplars are extracted using $k=9$.

\begin{figure}[H]
\center
\includegraphics[width=1\textwidth,height=0.4\textwidth]{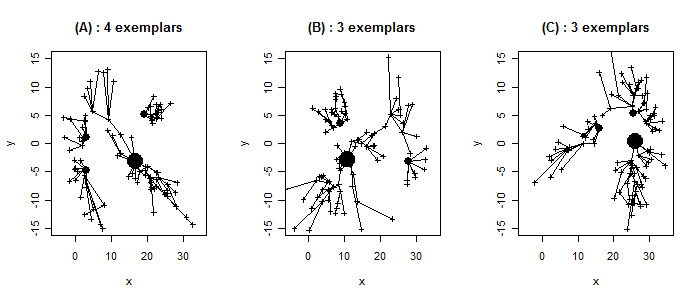}
\caption{
Optimal Scale Factor and Exemplars : The exemplars are extracted from the datasets (A), (B), and (C) of Figure~\ref{StandardExemples}. The optimal scale factors are respectively 9, 7 and 10. The numbers of exemplars (filled circles) are respectively 4, 3, and 3, larger filled circles show the standards.}
\label{optimumExemplars}
\end{figure}

Figure~\ref{optimumExemplars} displays the exemplars obtained with optimal scale factor from the datasets ((A), (B), and (C) on the Figure~\ref{StandardExemples}).
The random datasets have 100 samples ($n=100$).
$k_{optimum}$ is respectively equal to 9, 7 and 10.
The filled circles display the exemplars 
and a larger filled circle shows the standard.

\section{Applications}

This section presents applications of our method in two typical and very different contexts. The first application consists in extracting exemplars from a binary image database and building the graph of exemplars of this database. The second application present an analysis of the co-authoring in a research team by extracting exemplar authors and exhibiting the implicit structure.

\subsection{Extraction of exemplars from a set of binary images}

In this first application we consider a set of binary images contained in a database. The goal is to extract exemplar images from this database. The interest could be providing a set of resuming images or distinguishing subsets of images according to their content. The database is presented in the Table~\ref{Tab:imagettesScores}.\\
In a first we construct the relation matrix by using the Asymetric Haussdorff Distance. Classical methods of clustering have to work with \emph{symmetric} distance. They are inapplicable when distance from an image A to image B is not equal to distance from image B to image A. As we wrote at the beginning of this paper, the symmetry property is not required in our method.\\
\begin{table}[H]
\begin{center}
\begin{tabular}{|l|l|r|}
  \hline
Image & filename & score \\ 
  \hline
\includegraphics[width=1.00cm]{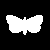} & \tt Butterfly-a004 & 12.16 \\ 
\includegraphics[width=1.00cm]{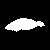} & \tt Fish-a023 & 10.22 \\ 
\includegraphics[width=1.00cm]{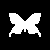} & \tt Butterfly-a029 & 9.34 \\ 
\includegraphics[width=1.00cm]{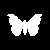} & \tt Butterfly-a011 & 9.22 \\ 
\includegraphics[width=1.00cm]{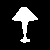} & \tt Lamp012 & 8.41 \\ 
\includegraphics[width=1.00cm]{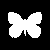} & \tt Butterfly-a009 & 8.06 \\ 
\includegraphics[width=1.00cm]{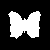} & \tt Butterfly-a014 & 7.78 \\ 
\includegraphics[width=1.00cm]{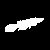} & \tt Fish-a019 & 7.69 \\ 
   \hline
   \end{tabular}\begin{tabular}{|l|l|r|}
  \hline
Image & filename & score \\ 
  \hline
\includegraphics[width=1.00cm]{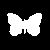} & \tt Butterfly-a001 & 7.62 \\ 
\includegraphics[width=1.00cm]{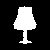} & \tt Lamp015 & 7.38 \\ 
\includegraphics[width=1.00cm]{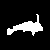} & \tt Fish-a018 & 7.25 \\ 
\includegraphics[width=1.00cm]{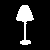} & \tt Lamp010 & 6.88 \\ 
\includegraphics[width=1.00cm]{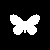} & \tt Butterfly-a028 & 5.34 \\ 
\includegraphics[width=1.00cm]{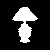} & \tt Lamp013 & 5.22 \\ 
\includegraphics[width=1.00cm]{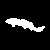} & \tt Fish-a030 & 3.94 \\ 
\includegraphics[width=1.00cm]{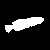} & \tt Fish-a035 & 3.50 \\ 
   \hline
   \end{tabular}
   \end{center}
\caption{Binary images sorted by decreasing score}
\label{Tab:imagettesScores}
\end{table}

\begin{figure}[H]
\begin{center}
\fbox{\includegraphics[width=\textwidth]{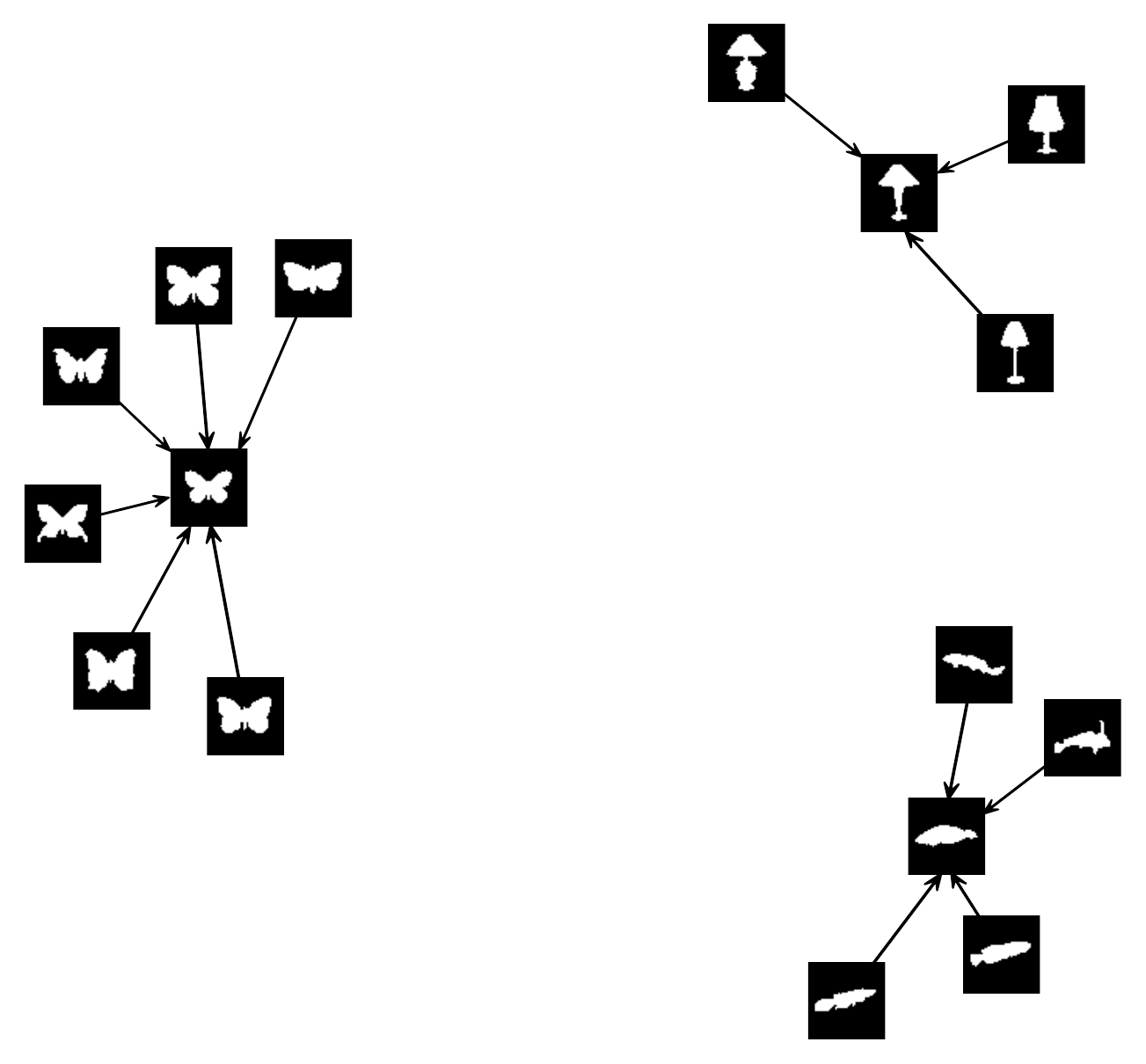}}\\
\end{center}
\caption{Network of the binary images where each image is connected to one exemplar. This directed graph exhibits three connected components forming three clusters coinciding with the content of images}
\label{Fig:IM_1}
\end{figure}

Firstly, we compute the score of each image of the database. The table \ref{Tab:imagettesScores} represents the images sorting by decreasing order of scores. Secondly we build the associated directed graph presented in Figure~\ref{Fig:IM_1} and representing the exemplars network (with a scale factor of $4$). This graph show how the dataset is structured. We can observe that the connected components of this graph are grouping image according to the object they represent. The three images of the Table~\ref{Tab:imagettesExemplars}  are the exemplars of this dataset and provide a good summary of the whole dataset.
\begin{table}[H]
\begin{center}
\begin{tabular}{ccc}
\includegraphics[width=0.1\textwidth]{Butterfly-a028} &
\includegraphics[width=0.1\textwidth]{Lamp012} &
\includegraphics[width=0.1\textwidth]{Fish-a023}
\end{tabular}
\end{center}
\label{Tab:imagettesExemplars}
\caption{Exemplars extracted from the set of binary images of the Table~\ref{Tab:imagettesScores}}.
\end{table}

\subsection{Exploration of co-authoring network}

The second application of our concept deals with publication data inside a laboratory, a research team or any other group of researchers. \\
Co-authoring informations can be considered as relational data (\cite{Mcgovern2003}, \cite{Neville2003}). In this work, we consider that the value of the relation from a researcher named Alice to a researcher named Bob is computed as the sum for each common publication of the product of the number of coauthor on the publication and the number of publication of Alice. This relation is not symmetric. In fact, generally, Alice can be the "preferred" co-author of Bob, but Bob is not necessarily the "preferred" co-author of Alice. This valued relation characterizes the "quality" of links between the  members and takes account of their publication activity.\\

The dataset we used is the set of publications of the CReSTIC Laboratory (University of Reims, France) \cite{Benassarou2010}. This informations are extracted from the web site of the laboratory and have been anonymized. \\

\begin{figure}[H]
\center
\fbox{\includegraphics[width=\textwidth]{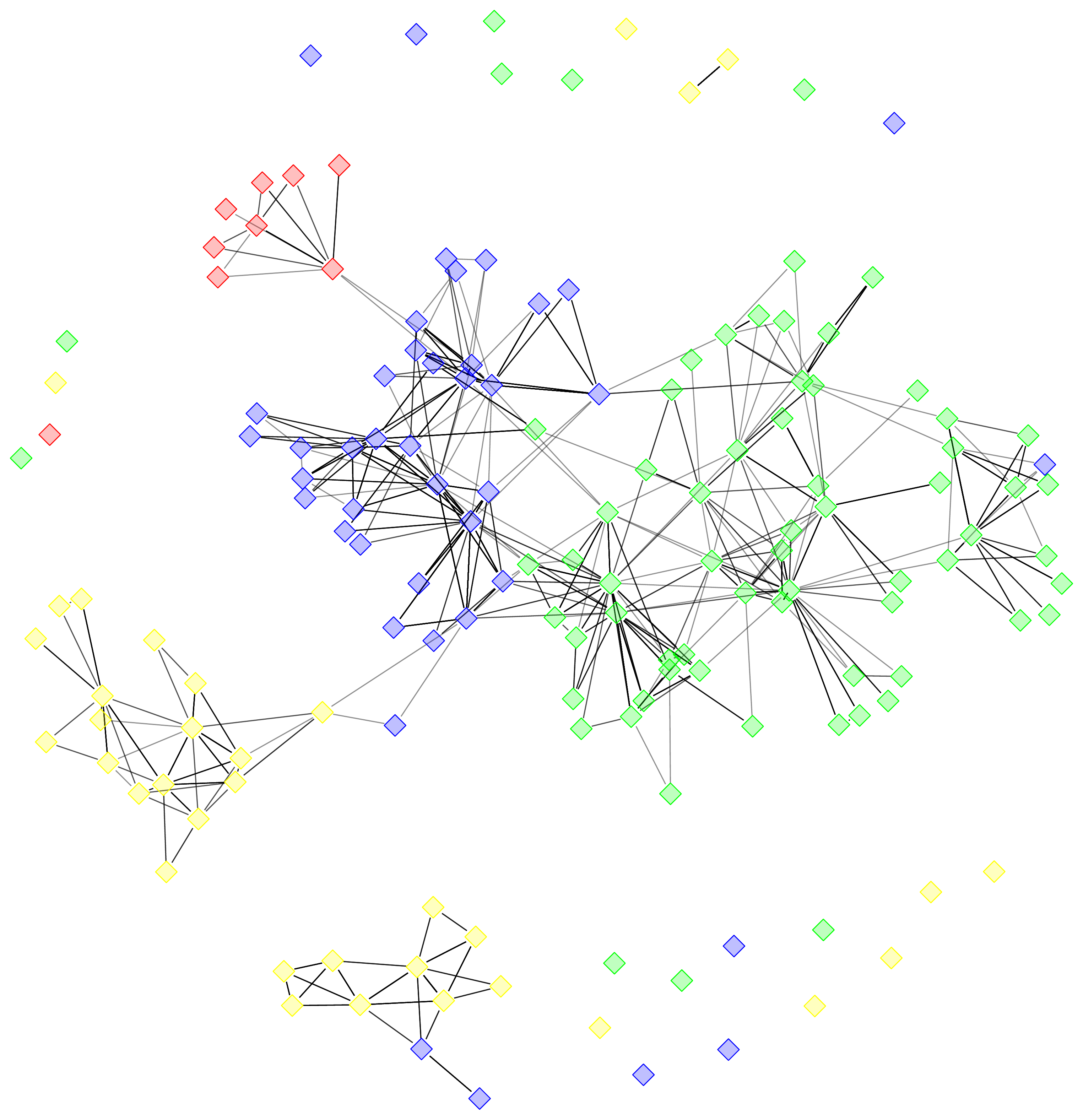}}
\caption{Digraph of the co-authoring in a laboratory. Each vertex is one researcher and each edge corresponds to one common publication.}
\label{Fig:CO1}
\end{figure}

The graph of the Figure~\ref{Fig:CO1} represents this dataset. Each node is a lab member and each edge between two members represents one common publication. Different colors are used to represents the different teams that compose the laboratory (but this information is not used in the computation of the exemplars). Therefore the scale factor is not used in this application because the size of the neighborhood is implicitly fixed in the dataset (according to the number of co-author of each member of the team).\\

\begin{figure}[H]
\center
\fbox{\includegraphics[width=\textwidth]{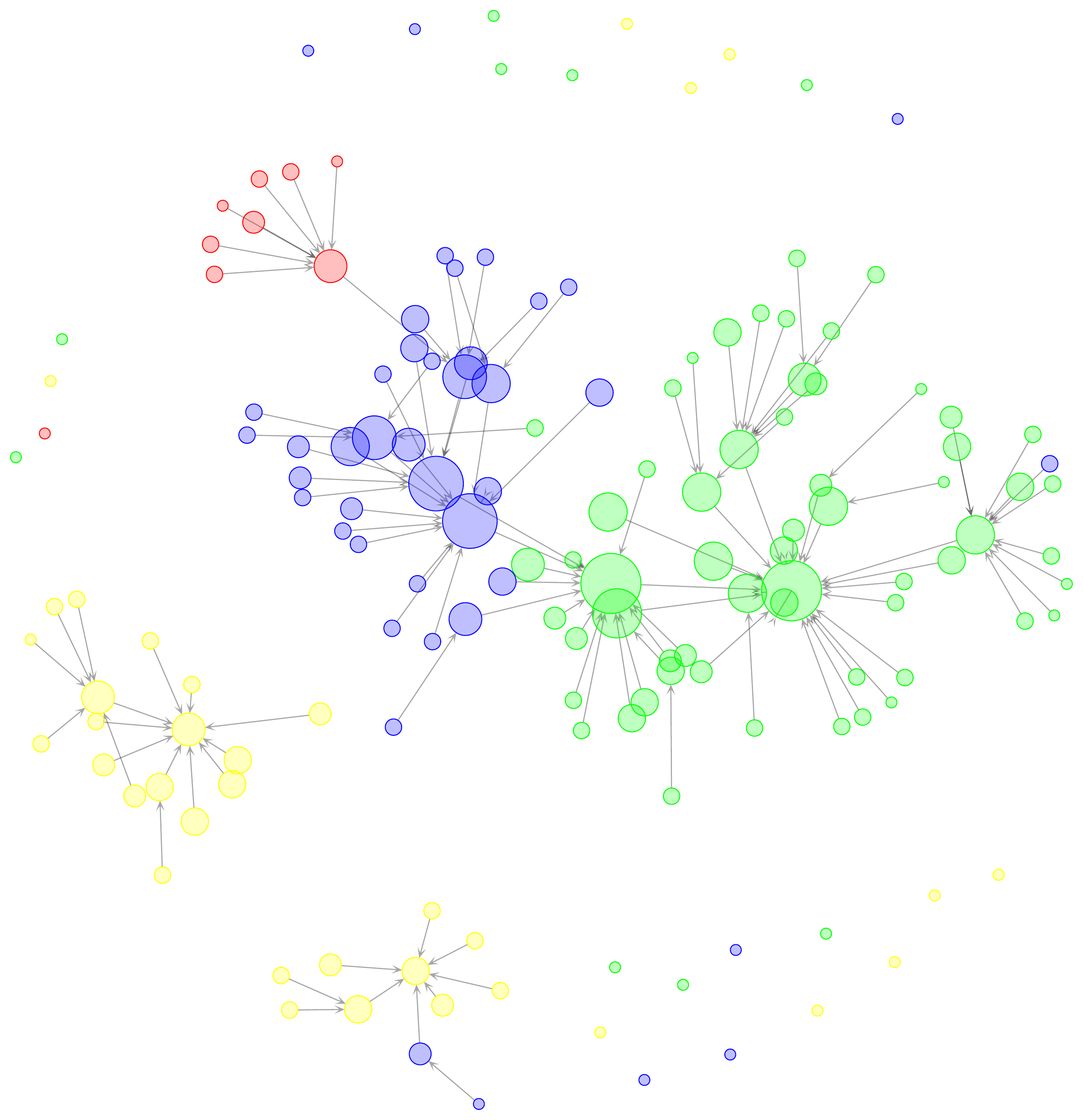}}
\caption{Representative Network extracted from data of Figure~\ref{Fig:CO1}. The higher is the score of one researcher, the higher is the diameter of its vertex in the graph. In this graph, each edge is the link of one researcher to its exemplar.}
\label{Fig:CO2}
\end{figure}

\begin{figure}[H]
\center
\fbox{\includegraphics[width=\textwidth]{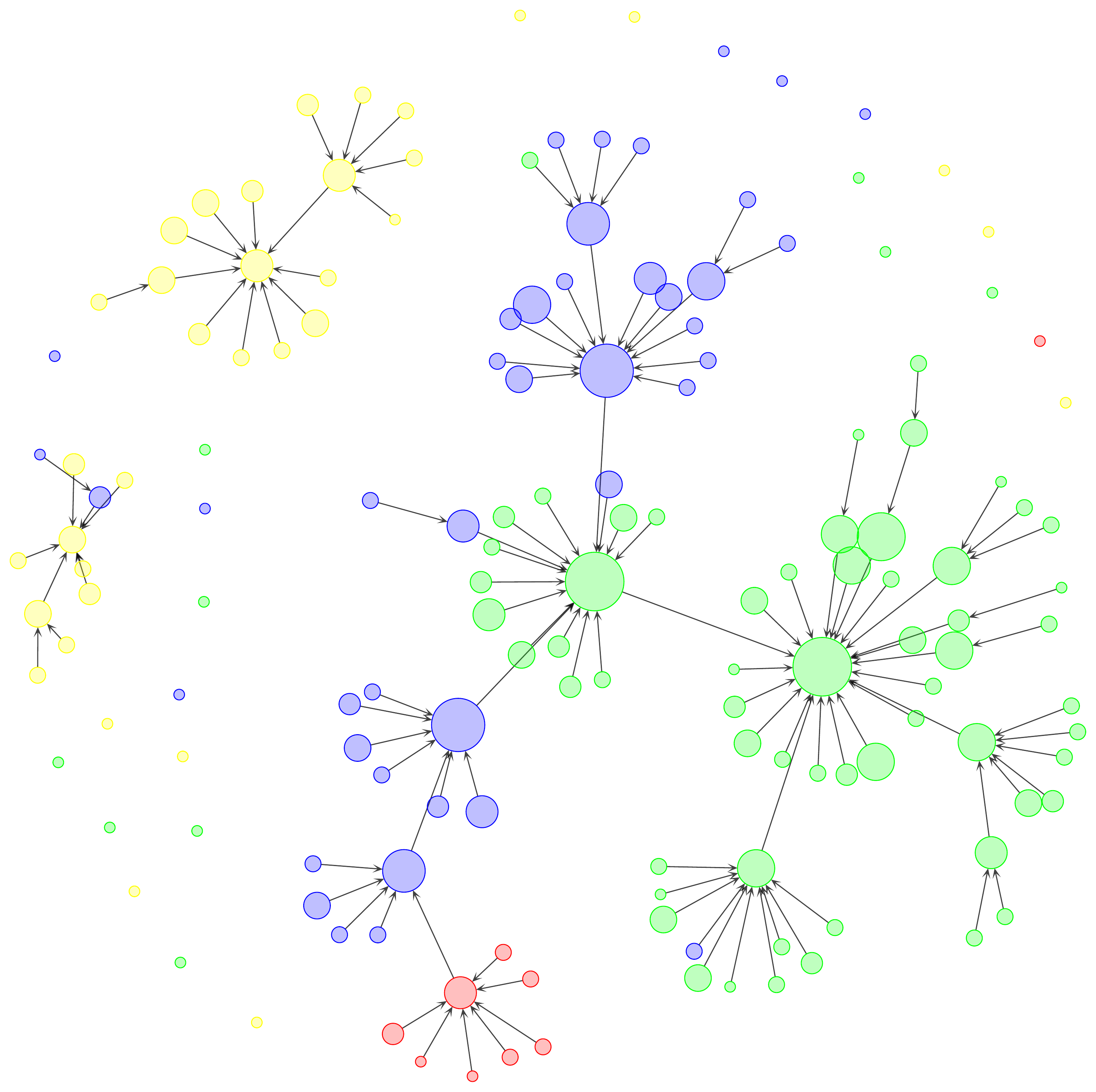}}
\caption{Network extracted from data of Figure~\ref{Fig:CO1} after rearranging the vertex positions (to increase the readability)}
\label{Fig:CO3}
\end{figure}

After computing the scores, we built the exemplars graph represented on the Figure~\ref{Fig:CO2}. The size of the node is proportional to its score. This graph is displayed using the same position for the nodes. In the Figure~\ref{Fig:CO3}, the nodes are rearranged to propose a clearer visualization.


The graphs presented in the Figures \ref{Fig:CO1},~\ref{Fig:CO2} and \ref{Fig:CO3} show several interests of our method. The first interests is the simplification of the graph of the Figure \ref{Fig:CO1}. When the numbers of vertices and edges are growing the graph becomes more unreadable. For big data, resuming and simplifying is a necessary task.\\
The second interest is to exhibit such a sub-structure of the team (this task is called community detection in a network \cite{citeulike:3070598}). The Figures \ref{Fig:CO2} and \ref{Fig:CO3} show how groups are connected, and which members are the most representative. The exemplars members are connecting the others and can be viewed as natural leaders (or natural mentors) according to their publications and their co-authors. It emphasizes the important (critical) position of some members in a research team.\\
Incidentally, we can observe that the resulting clustering obtained by partitioning the graph in connected components is a little bit different of the real partitioning in sub-groups (represented by the different colors)

\section{Conclusion}

In the framework of data mining, this paper describes a new way 
for extracting exemplars from a relational dataset.
The method we propose is based on a pairwise comparison
assuming a coarse relation on the dataset.
This approach is particularly adapted 
when no distance is available or meaningful in the data domain.
Moreover the coarse relation between data 
does not need symmetry or transitivity properties.
Thus the method is useful for any kinds of relational data. \\
An aggregated score is defined from these pairwise comparisons.
The paper defines the standard
which is the sample with the highest score.
Simulations show the robustness of the standard against outliers
and the stability of the standard when resampling the dataset.
Thus these results confirm the standard
as a robust location estimator.
Moreover the aggregated score is used to extract exemplars which are real objects.
Then our approach of location estimator avoids 
the drawbacks of average objects which are meaningless
when processing qualitative data.\\
Using a score based on the pairwise comparison, 
we define the k nearest neighbors of each datum.
This approach permits us to extract exemplars depending on this k value.
We state that the number of local exemplars decreases from n to 1
(n is the number of data samples)
when k value increases from 1 to n.
Thus k is considered as a scale factor.
The method we propose allows us to explore the dataset through different scales.
We can adjust the k value for extracting a reduced number of exemplars.
An automated approach is proposed to determine an optimal number of exemplars.\\
On top of the extraction of exemplars, 
the method proposes to design a network.
The paper shows that the network is reconfigured 
when the scale factor changes. 
The network eases the explanation of the exemplar roles in the dataset.
When the scale factor increases, some exemplars could disappear
keeping the most important ones 
(i.e. the exemplars which are important nodes for connecting some data).\\

In future works we propose to use the fuzzy set theory as in \cite{Blanchard2010} to generalize our framework in the case of fuzzy relation, when ranking data is not easy.\\ The major way we would to explore is the area of Social Network Analysis. We are convinced that our concept of \emph{exemplar} could be a significant tool for extracting leaders or mentors in social network and improve recommendation systems. Our concept of degree of representativeness should be compared to the different definitions of \emph{centrality} in a network \cite{Pfeiffer2010}.

\bibliography{biblio}

\end{document}